\documentclass{article}

\usepackage{microtype}
\usepackage{graphicx}
\usepackage{subfigure}
\usepackage{booktabs} % for professional tables
\usepackage{multirow}
\usepackage{hyperref}

\usepackage[accepted]{icml2025}

\usepackage{amsmath}
\usepackage{amssymb}
\usepackage{mathtools}
\usepackage{amsthm}

\usepackage[capitalize,noabbrev]{cleveref}

\theoremstyle{plain}

\theoremstyle{definition}

\theoremstyle{remark}

\usepackage[textsize=tiny]{todonotes}

\NewDocumentCommand{\yafu}
{ mO{} }{\textcolor{blue}{\textsuperscript{\textit{yafu}}\textsf{\textbf{\small[#1]}}}}

\icmltitlerunning{}

\begin{document}

\twocolumn[
\icmltitle{New Skills or Sharper Primitives? A Probabilistic Perspective on the Emergence of Reasoning in RLVR}



\icmlsetsymbol{equal}{*}

\begin{icmlauthorlist}
\icmlauthor{Zhilin Wang}{ustc,shai}
\icmlauthor{Yafu Li}{shai,cuhk}
\icmlauthor{Shunkai Zhang}{pku}
\icmlauthor{Zhi Wang}{nju}
\icmlauthor{Xiaoye Qu}{shai}
\icmlauthor{Haoran Zhang}{shai,sjtu}
\icmlauthor{Yu Cheng}{cuhk}
\end{icmlauthorlist}

\icmlaffiliation{ustc}{University of Science and Technology of China}
\icmlaffiliation{cuhk}{The Chinese University of Hong Kong}
\icmlaffiliation{pku}{Peking University}
\icmlaffiliation{nju}{Nanjing University}
\icmlaffiliation{shai}{Shanghai AI Laboratory}
\icmlaffiliation{sjtu}{Shanghai Jiao Tong University}

\icmlcorrespondingauthor{Yafu Li}{yafuly@gmail.com}
\icmlcorrespondingauthor{Yu Cheng}{chengyu@cse.cuhk.edu.hk}
%\icmlcorrespondingauthor{}{@www.uk}

\icmlkeywords{Machine Learning, ICML}

\vskip 0.3in
]




\begin{abstract}
Whether Reinforcement Learning with Verifiable Rewards (RLVR) endows Large Language Models (LLMs) with new capabilities or merely elicits latent traces remains a central debate. In this work, we align with the former view, proposing a probabilistic framework where capability is defined by instance-level solvability. We hypothesize that the emergence of complex reasoning can be driven by sharpening atomic step probabilities, which enables models to overcome the exponential decay of success rates inherent in multi-step reasoning chains. Utilizing the Algebrarium framework, we train models exclusively on single-step operations and evaluate their performance on unseen multi-step tasks. Our empirical results confirm that: (1) RLVR incentivizes the exploration of previously inaccessible solution paths by amplifying the model’s existing skills; (2) composite performance is strictly governed by the joint probability of atomic steps, evidenced by high Pearson correlation coefficients ($\rho \in [0.69, 0.96]$); and (3) RLVR, acting as a global optimizer, can cause specific skills to be sacrificed to maximize aggregate reward. Our work offers a novel explanation for emergent abilities in RLVR, suggesting that the iterative optimization of solvable problems enables models to develop the capabilities to tackle previously unsolvable scenarios.

%These findings provide a theoretical and empirical foundation for understanding RL as an optimizer of compositional potential more than a creator of novel knowledge.
\end{abstract}

\printAffiliationsAndNotice % 

\section{Introduction}

While the Reinforcement Learning with Verifiable Rewards (RLVR,~\citealp{Shao2024DeepSeekMathPT}) paradigm has achieved impressive success~\cite{DBLP:journals/corr/abs-1910-07113,GDM2025Gemini2P}, a central question is whether it endows Large Language Models (LLMs) with genuinely new capabilities or merely elicits traces already latent within the models themselves. This question stands at the forefront of current research. Some studies argue for the former, positing that RLVR implants new compositional skills absent in the base model~\cite{yuan2025fxgxfgxllms,cheng2025atomiccompositereinforcementlearning}; conversely, other works maintain the latter, suggesting that RLVR acts primarily as a mechanism to unlock pre-existing potentials~\cite{Yue2025DoesRL,Gandhi2025CognitiveBT,Liu2025DrGRPO,Zhao2025EchoChamber}. 

In this work, we align with the former perspective, anchoring our argument in a rigorous probabilistic definition of capability: if a model fails to produce a solution despite exhaustive sampling, it is fundamentally devoid of the corresponding capability. Such complex problems often necessitate sequential reasoning, where success hinges on the correctness of each intermediate step~\cite{wei2023chainofthoughtpromptingelicitsreasoning}. While the base model may possess the competence to solve individual atomic steps, the cumulative probability of executing a perfect chain is negligible, rendering the complex task practically unsolvable~\cite{dziri2023faithfatelimitstransformers}. 
By significantly improving the accuracy of each individual step, RLVR prevents errors from accumulating in long chains. This allows the model to reliably complete complex reasoning paths that were previously too unlikely to succeed. Thus, while RLVR may not impart novel knowledge~\cite{zhou2023limaalignment}, it nonetheless unlocks a genuinely new capability through reliable composition.

Our work begins with a comprehensive theoretical analysis of \textit{Pass@k} across the entire dataset, decomposing aggregate metrics to extract the specific success rate for each individual problem. We propose using this instance-level solvability as the intrinsic measure of capability. 
By scrutinizing the success rates of complex tasks, we uncover a theoretical bottleneck: the \textit{Multiplicative Barrier}. We demonstrate that the vanishing solvability for reasoning-heavy problems is a mathematically inevitable consequence of exponential decay, where the joint probability of a solution collapses as the product of atomic step probabilities.
% ($P_\theta(q) \approx \prod P_\theta(s_j)$). 
Consequently, we frame RLVR-induced emergence as a phase transition where the sharpening of atomic probabilities enables the model to pierce through this multiplicative ceiling.
Furthermore, we characterize \textit{Capability Erosion} as a side effect of optimizing for expected utility. Since the RLVR objective optimizes from a global distributional perspective, the resulting gradient updates do not inherently preserve instance-level stability, potentially leading to performance degradation on specific instances.

To rigorously validate these theoretical assumptions, we conduct a controlled experimental study using the Algebrarium framework.   
% (detailed in Appendix~\ref{app:algebrarium}). 
We design a strict ``Atomic-Composite Split" protocol to isolate the mechanism of emergence: models are trained exclusively on decoupled atomic primitives but evaluated on their ability to solve unseen composite problems. 
Our empirical analysis yields four critical insights into the nature of capability emergence:

\begin{itemize}
    \item \textbf{Validation of \textit{Pass@k} Dynamics:} 
    % \textcolor{black}{By formulating \textit{Pass@k} as a function of instance-level correctness probabilities, we achieve a high-precision alignment ($MSE \leq 10^{-4}$) between our theoretical curves and empirical observations.}  %\yafu{not clear}
    We derive \textit{Pass@k} as an explicit function of per-instance correctness probabilities and show that the resulting curves closely match empirical measurements.
    % ($MSE \leq 10^{-4}$).
    % \item \textbf{Emergence from the Null Regime:} We visualize the capability emergence, showing how RLVR resurrects tasks from \textbf{the Null Set} to \textbf{the Feasible Set}.  
    \item \textbf{Operational definition of emergence:} Using instance-level solvability thresholds, we visualize a phase transition in which RLVR moves problems from the \textbf{Null Set} to the \textbf{Feasible Set}.
    % \item \textbf{Verification of the Multiplicative Barrier:} We empirically verify the exponential decay of reasoning success ($P \propto p^N$) and demonstrate a strong statistical dependency between the joint atomic accuracy and composite task performance, evidenced by high Pearson correlation coefficients ($\rho \in [0.69, 0.96]$).
    \item \textbf{Mechanism via the multiplicative barrier:} We verify that multi-step reasoning success decays exponentially with reasoning depth ($P \propto p^N$) and that composite performance is strongly predicted by the joint probability of atomic steps (Pearson $\rho \in [0.69, 0.96]$).
    % \item \textbf{Evidence of Capability Erosion:} \textcolor{black}{Even within atomic tasks, we observe that RLVR can lead to the compromise of specific skills in favor of maximizing the aggregate reward. Consequently, while RLVR facilitates the acquisition of new capabilities, it may simultaneously induce regression in others.}
    \item \textbf{Trade-off via capability erosion:} We document that optimizing expected reward can improve aggregate performance while degrading specific (including atomic) instances, highlighting a redistribution of capability rather than uniform gains.
\end{itemize}

In conclusion, our work offers a novel explanation for emergent abilities in RLVR, suggesting that the iterative optimization of solvable problems enables models to develop the capabilities to tackle previously unsolvable scenarios.

\section{Related Work}

\paragraph{Reinforcement Learning with Verifiable Rewards (RLVR).}
The RLVR paradigm has become pivotal for enhancing LLM reasoning abilities~\cite{ElKishky2024OpenAIOS,GDM2025Gemini2P}. 
These algorithms update the policy based on the relative advantage of sampled paths, reinforcing trajectories that outperform the group advantage~\citep{DeepSeekAI2025DeepSeekR1IR, Shao2024DeepSeekMathPT,yu2025dapoopensourcellmreinforcement,zheng2025groupsequencepolicyoptimization}. However, the efficacy of this sampling-based optimization has sparked a fundamental debate: does this process merely surface latent traces, or does it genuinely endow the model with new capabilities~\cite{Yue2025DoesRL}?

\paragraph{Perspective I: RLVR merely elicits latent traces.}
A growing body of research suggests that RLVR amplifies existing competencies rather than injecting new capabilities~\citep{Zhao2025EchoChamber,wen2025reinforcement}. 
Empirically, RLVR training is observed to reduce entropy~\citep{cui2025entropymechanismreinforcementlearning,wang2025beyond,jiang2025rethinkingentropyregularizationlarge,yan2025learningreasonoffpolicyguidance} and often degrade \textit{Pass@k} performance compared to baselines, indicating a contraction of exploration space rather than capability expansion~\citep{Yue2025DoesRL,zhu2025surprising,he2025rewarding}.
Theoretically, this limitation stems from the dependence of on-policy algorithms on spontaneous generation: if a correct solution cannot be sampled, the model receives no reward signals~\citep{zhan2025exgrpolearningreasonexperience}. Thus, RLVR is viewed merely as a mechanism to elicit latent potentials already encoded during pre-training.

\paragraph{Perspective II: RLVR implants new capabilities.}
Conversely, some recent studies challenge the elicitation hypothesis, arguing that RLVR can induce novel, complex composite capabilities~\citep{yuan2025fxgxfgxllms,cheng2025atomiccompositereinforcementlearning}. 
To control for confounding environmental factors, researchers employ synthetic tasks such as the Countdown problem or design specific environments to disentangle atomic skills from composite reasoning~\cite{park2025doesrlposttraininginduce}. 
Contrary to the first perspective, these works observe a substantial improvement in \textit{Pass@k} after training, suggesting that RLVR facilitates the induction of novel reasoning paths rather than the mere elicitation of latent ones. 
However, the underlying mechanism driving this emergence remains underexplored.

Our work provides both theoretical and experimental support for the second perspective, demonstrating that RLVR induces novel capabilities by optimizing solvable atomic sub-tasks to construct complex reasoning paths previously absent from the base model.
\section{Understanding the \textit{Pass@k} Curve}

\begin{figure*}[!t]
    \centering
    \includegraphics[width=1\linewidth]{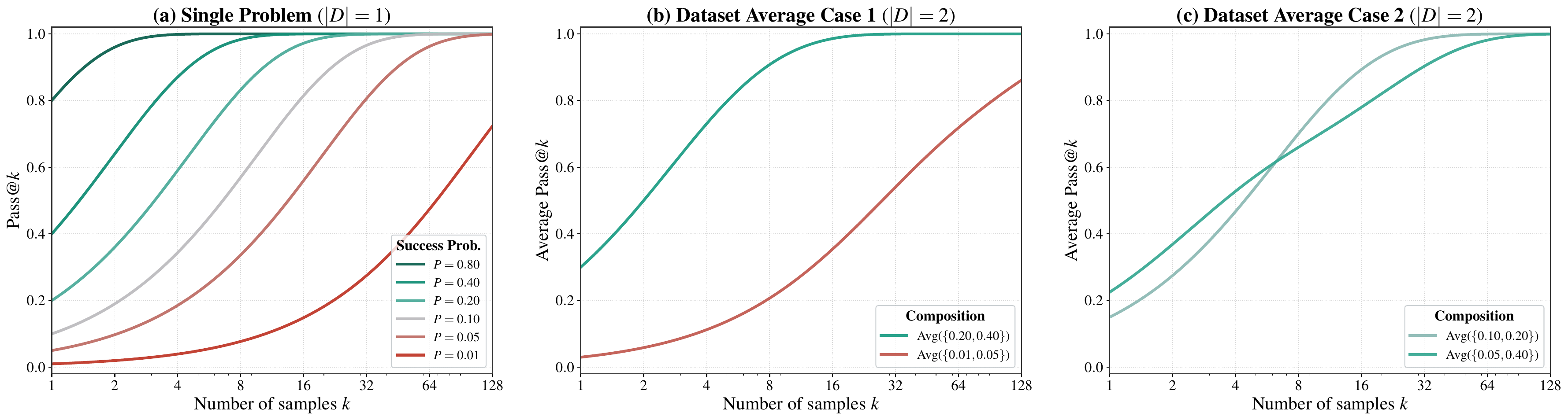}
    \caption{Theoretical visualization of $Pass@k$. (a) Functional curves for a single problem under varying correctness probabilities. (b) and (c) Composite average curves for a simplified dataset ($|\mathcal{D}|=2$), showing the aggregation from instances to the dataset-level metric. %\yafu{put in the previous page}
    }
    \label{fig:pass@k.theory}
\end{figure*}

\subsection{Theoretical Analysis}
\label{sec:theory_passk}

Given a dataset $\mathcal{D} = \{q_i\}_{i=1}^N$ consisting of $N$ problems, the \textit{Pass@k} metric measures the proportion of problems successfully solved by at least one solution when $k$ candidate solutions are generated per problem:
\begin{align}
\textit{Pass@k} = \frac{1}{|\mathcal{D}|} \sum_{i=1}^{|\mathcal{D}|} \mathbb{I}(\text{solved}_i \geq 1), 
\end{align}
where $\mathbb{I}(\cdot)$ is the indicator function, and $\text{solved}_i \in \{0, 1, \dots, k\}$ represents the count of successful samples for the $i$-th problem out of $k$ independent generations.

To formalize this, let $\pi_\theta(y|q)$ denote the policy of the model generating a solution $y$ for a query $q \in \mathcal{Q}$. Let $V(y, q) \in \{0, 1\}$ be the correctness verification function. The success probability for a specific task $q$ is defined as the expectation of correctness over the model's generation distribution:
\begin{align}
    P_\theta(q) = \mathbb{E}_{y \sim \pi_\theta(\cdot|q)} \left[ V(y, q) \right]
    \label{eq:expect}
\end{align}

First, consider a single problem $q$. Assuming $k$ samples are generated independently, the probability that all $k$ samples fail is $(1 - P_\theta(q))^k$. Consequently, the expected \textit{Pass@k} for this single instance is:
\begin{align}
    E[\textit{Pass@k}(q)] = 1 - (1 - P_\theta(q))^k
\end{align}

Generalizing this to the entire dataset $\mathcal{D}$, the overall $pass@k$ is defined as the arithmetic mean of the expected pass rates across all $N$ instances:
\begin{align}
     E[\textit{Pass@k}] = \frac{1}{N} \sum_{i=1}^{N} \left[ 1 - (1 - P_\theta(q_i))^k \right]
\end{align}

By analyzing the behavior of the \textit{Pass@k} metric, we can disentangle distinct dimensions of model performance.

\subsection{From Theory to Estimation}

To intuitively illustrate this theoretical framework, Figure~\ref{fig:pass@k.theory} presents two complementary views. Figure~\ref{fig:pass@k.theory}(a) plots the functional curves for a single problem under varying correctness probabilities, demonstrating how the curve's shape is strictly determined by the specific correctness rate. In contrast, Figure~\ref{fig:pass@k.theory}(b) and (c) consider simplified datasets where $|\mathcal{D}|=2$. They visualize the composite average curve formed by aggregating two distinct problem instances, highlighting the theoretical composition process from individual problems to the dataset-level metric.

\textbf{Post-RLVR Degradation in High-\(k\) Sampling.}
The phenomenon of decreasing \textit{Pass@k} after RLVR, as noted in prior work~\cite{Yue2025DoesRL}, can be explained using Figure~\ref{fig:pass@k.theory}. Specifically, post-RLVR models in their work tend to increase \textit{Pass@k} at low $k$ values while decreasing it at high $k$. This suggests that RLVR polarizes their model's success probabilities across the task distribution. It drives the correctness probability of certain problems towards certainty, while inadvertently reducing the probability of solving others. Consequently, this polarization improves \textit{Pass@1},  but harms the broader coverage measured by high-$k$ metrics.

\section{Capability as Probabilistic States}
\label{sec:capability_determination}

To rigorously quantify model capability, we move beyond aggregate metrics and analyze the pointwise convergence of $P_\theta(q)$. We posit that for a given model $\pi_\theta$, the task space $\mathcal{Q}$ can be partitioned based on the solvability of tasks under a computational budget constraint.

Based on the magnitude of $P_\theta(q)$, we partition the task space $\mathcal{Q}$ into distinct capability regions. We introduce two critical thresholds: an infinitesimal bound $\epsilon \to 0^+$ representing statistical impossibility, and a feasibility lower bound $\delta \in (0, 1]$ representing reliable solvability.

\paragraph{Definition 1: The Null Set ($\epsilon$-Incapability).}
A task $q$ belongs to the \textit{Null Set} $\mathcal{N}_\theta^\epsilon$ if the probability of generating a correct solution is asymptotically negligible. This state represents a topological disconnection between the problem and the model's reachable hypothesis space. Formally:
\begin{equation}
    \mathcal{N}_\theta^\epsilon = \left\{ q \in \mathcal{Q} \mid P_\theta(q) < \epsilon \right\}
\end{equation}
Empirically, we determine $\epsilon$ using the statistical \textit{Rule of Three}. If zero successes are observed in a large-sample regime (e.g., $K_{\text{large}}=128$), the upper bound of the 95\% confidence interval for the true success rate is approximately $3/K_{\text{large}}$. Thus, we set $\epsilon \approx 0.023$. A task in this set implies that correctness is effectively a zero-probability event under the current model parameters.

\paragraph{Definition 2: The Feasible Set ($\delta$-Capability).}
Conversely, a task is considered \textit{Feasible} if the success probability is bounded away from zero by a sufficient margin $\delta$. This ensures that the task can be solved reliably within a finite compute budget. We define the Feasible Set $\mathcal{F}_\theta^\delta$ as:
\begin{equation}
    \mathcal{F}_\theta^\delta = \left\{ q \in \mathcal{Q} \mid P_\theta(q) \geq \delta \right\}
\end{equation}
Here, $\delta$ is calibrated to a minimum compute budget $K_{\min}$ (e.g., standard parallel inference where $K_{\min}=8$). By setting $\delta = 1/K_{\min} = 0.125$, we guarantee that the expected number of correct solutions is at least 1:
\begin{equation}
    K_{\min} \cdot P_\theta(q) \geq 1
\end{equation}
This condition distinguishes stable capability from stochastic outliers (``lucky guesses''), ensuring that the model possesses the intrinsic features required to solve $q$.

\paragraph{The Stochastic Gap.}
Tasks falling in the interval $\mathcal{Q} \setminus (\mathcal{N}_\theta^\epsilon \cup \mathcal{F}_\theta^\delta)$, where $\epsilon \le P_\theta(q) < \delta$, are classified as \textit{Transitional States}. These tasks exhibit high variance and represent the frontier of the model's learning process. For the purpose of robust capability analysis, we focus our subsequent discussion on the phase transition between the limit of impossibility ($\mathcal{N}_\theta^\epsilon$) and the state of reliable capability ($\mathcal{F}_\theta^\delta$).

\section{Gain and Loss of Capability}
\label{sec:emergence_and_erosion}

In this section, we analyze the dual impact of RLVR optimization on model capabilities, exploring two simultaneous phenomena driven by the model's finite capacity. First, we define \textit{Compositional Emergence}, illustrating how the linear amplification of atomic steps triggers a geometric phase transition, rendering previously intractable complex tasks solvable. Then, we examine \textit{Capability Erosion}, demonstrating that RLVR optimization for global rewards can inadvertently degrade specific skills.

\subsection{The Multiplicative Barrier in Complex Reasoning}

Consider a complex problem $q$ that can be decomposed into a sequence of $M$ sub-problems, denoted as $\mathcal{S} = \{s_1, s_2, \dots, s_M\}$. For the model to successfully solve $q$, it must correctly execute each sub-problem $s_j$ in the chain.

Under the assumption that the correctness of each step is independent, the overall success probability $P_\theta(q)$ is the product of the success probabilities of the individual sub-problems:
\begin{equation}
    P_\theta(q) \approx \prod_{j=1}^{M} P_\theta(s_j)
\end{equation}
This multiplicative nature creates a severe barrier for deep reasoning. Even if the model exhibits reliable capability for every atomic sub-problem (e.g., $P_\theta(s_j)=0.3$), the joint probability decays exponentially with the chain length $M$.

For a problem requiring $M=5$ such steps, the cumulative success rate collapses to:%\yafu{use \$\$}
$$P_\theta(q) \approx 0.3^5 \approx 0.0024$$
Crucially, this value falls an order of magnitude below the statistical existence threshold $\epsilon \approx 0.023$ (the Null Set boundary). Then, while the model is locally capable of every step, it appears globally incapable of solving the whole problem $q$. Empirically, this results in zero successes ($Pass@128 = 0$), masking the model's underlying potential.

\subsection{Emergence through Probability Amplification}

RL can operate by optimizing these atomic primitives where reward signals are dense. By sufficiently ``sharpening'' the atomic probabilities, RLVR triggers a multiplicative phase transition at the compositional level.
We formalize this emergence as the crossing of the feasibility boundary:
\begin{equation*}
    \underbrace{0.3^5 \approx 0.0024}_{\substack{\text{Null State} \\ (P_\theta(q) < \epsilon)}} \xrightarrow[\text{Sharpening}]{\text{RL}} \underbrace{0.7^5 \approx 0.168}_{\substack{\text{Feasible State} \\ (P_\theta(q) \ge \delta)}}
\end{equation*}
Here, the new capability to solve $Q_{comp}$ emerges not from learning the complex task logic directly, but from the saturation of prerequisite atomic competencies. This resolves the central paradox: RLVR acts as an amplifier at the micro-level, which mathematically necessitates its appearance as a creator at the macro-level.

\begin{table*}[t]
    \centering
    \small
    \renewcommand{\arraystretch}{1.15}
    \setlength{\tabcolsep}{6pt}
    \caption{\textbf{Dataset Statistics and Domain Properties.} We evaluate four algebraic systems varying in cardinality and commutativity. The training set consists exclusively of atomic examples to enforce rule learning, while the test set comprises 800 samples stratified by reasoning depth (2--5 operations) to assess generalization.}
    \label{tab:experiment_overview}
    
    \begin{tabular}{l|cc|l|c|cccc|c}
        \toprule
        \multirow{2}{*}{\textbf{Domain}} & \multicolumn{2}{c|}{\textbf{Group Properties}} & \multirow{2}{*}{\textbf{Underlying Structure}} & \textbf{Train} & \multicolumn{4}{c|}{\textbf{Test (Stratified by Depth)}} & \multirow{2}{*}{\textbf{Total}} \\
        \cmidrule(lr){2-3} \cmidrule(lr){6-9}
         & Card. & Comm. & & \textbf{1-Ops} & \textbf{2-Ops} & \textbf{3-Ops} & \textbf{4-Ops} & \textbf{5-Ops} & \\
        \midrule
        Encrypted History & $\infty$ & Yes & Integers ($\mathbb{Z}$) & 3,200 & 50 & 50 & 50 & 50 & 200 \\
        Enigma System & Finite & Yes & Modular Product ($\mathbb{Z}_{26}^3$) & 3,200 & 50 & 50 & 50 & 50 & 200 \\
        Knitting System & $\infty$ & No & Free Group & 3,200 & 50 & 50 & 50 & 50 & 200 \\
        Rubik's Cube & Finite & No & Permutation Group & 3,200 & 50 & 50 & 50 & 50 & 200 \\
        \midrule
        \textbf{Total} & \multicolumn{3}{c|}{\textit{4 Distinct Algebraic Systems}} & \textbf{12,800} & \textbf{200} & \textbf{200} & \textbf{200} & \textbf{200} & \textbf{800} \\
        \bottomrule
    \end{tabular}
\end{table*}

\subsection{Capability Erosion under RLVR Optimization}
\label{theory:erosion}
While RLVR acts as an amplifier for emergence, this optimization is not a monotonic improvement across the entire task domain. Instead, it often triggers \textit{Capability Erosion}, a phenomenon where the mastery of certain problem instances comes at the expense of others. 

The RLVR objective is defined by the maximization of the expected reward over the task distribution $\mathcal{D}$:

\begin{equation}
    \max_\theta \mathcal{J}(\theta) = \frac{1}{N} \sum_{i=1}^{N} \mathbb{E}_{x \sim \pi_\theta(\cdot |q_i)} [R(x)]
\end{equation}

By maximizing mean reward, RLVR prioritizes the majority over the specific. This creates an inherent conflict: to master dominant tasks, the model may overwrite representations required for minority cases. As a result, training does not yield a uniform improvement; rather, it triggers a redistributive trade-off. Ultimately, the erosion of minority skills is the price of global improvement, showing that RL reshapes through destruction rather than simple accumulation.

\section{Experiment Setup}
\label{sec:experiment_setup}
\textcolor{black}{To isolate genuine logical deduction from memorized knowledge, we employ Algebrarium to generate synthetic symbolic reasoning tasks. This framework mitigates contamination by instantiating non-standard algebraic structures that do not exist in pre-training corpora. It provides a clean environment characterized by tunable complexity, deterministic answers, and a modular design that makes the reasoning process highly analyzable. By decoupling abstract definitions from concrete instances, Algebrarium ensures that performance reflects pure symbolic proficiency rather than data leakage. (Detailed in Appendix \ref{app:algebrarium}.)}
%\yafu{strengthen the motivation of Algebrarium, e.g., avoid irrelevant factors including contamination, controllability of difficulty, analyzable, with easy-to-verify answers.}
%We utilize Algebrarium, a framework designed to generate synthetic data for abstract algebra and symbolic reasoning, to evaluate reasoning capabilities in a controlled environment. By procedurally generating algebraic structures, we ensure that the evaluation isolates logical deduction from the retrieval of memorized knowledge found in pre-training corpora. Appendix~\ref{app:algebrarium} provides a detailed introduction to this framework.

\paragraph{Domains and Task Formulation.}
The core task is \textit{Algebraic Expression Calculation}. Given two elements $e_1, e_2 \in G$ from a specific group, the model must compute the canonical result of $e_1 \oplus e_2$ by applying group axioms (e.g., associativity, identity, inverse).
To test robustness across diverse algebraic properties, we configure four distinct domains defined by two axes: Cardinality (Finite vs. Infinite) and Commutativity (Abelian vs. Non-Abelian). 

\begin{table}[h]
\centering
\small
\caption{\textbf{Evaluation Framework.} The framework distinguishes between direct inference and decomposed binary operations. Note that all intermediate steps $d_j$ are defined as ground truth labels to allow for independent assessment of atomic operations ($P_\theta(s_j)$).}
\label{tab:binary_decomposition}
\begin{tabular}{@{}l c c@{}}
\toprule
\textbf{Mode} & \textbf{Structure (Binary Ops)} & \textbf{Metric} \\ \midrule

\textbf{Direct} & \multicolumn{1}{c}{$q \xrightarrow{\text{Infer}} \text{Result}$} & $\boldsymbol{P_\theta(q)}$ \\ 
\midrule

\textbf{Decomposed} & \multicolumn{1}{c}{\textit{Sequential Chain}} & \\
\quad Step 1 & $s_1: e_1 \oplus e_2 \to d_1$ & $P_\theta(s_1)$ \\
\quad \dots & \dots & \dots \\
\quad Step $j$ & $s_j: e_{k} \oplus d_{j-1} \to d_j$ & $P_\theta(s_j)$ \\
\quad Step $N$ & $s_N: e_{final} \oplus d_{N-1} \to \text{Res}$ & $P_\theta(s_N)$ \\ \midrule

\textbf{Definitions} & \multicolumn{2}{l}{\footnotesize $\bullet$ \textbf{Atomic:} $P_\theta(s_j)$ \quad $\bullet$ \textbf{Result:} $P_\theta(q)$} \\
& \multicolumn{2}{l}{\footnotesize $\bullet$ \textbf{Joint:} $\prod_{j=1}^{N} P_\theta(s_j)$} \\ 
\bottomrule
\end{tabular}
\end{table}

\textcolor{black}{
\textbf{Measurement and Decomposition.} 
As shown in Table~\ref{tab:binary_decomposition}, We quantify difficulty by operation count $N$ and compare direct inference $P_\theta(q)$ with decomposed binary steps $s_j$. Our analysis relates instance accuracy $P_\theta(q)$ to atomic step accuracy $P_\theta(s_j)$ and the joint probability $\prod_{j=1}^N P_\theta(s_j)$.
}

\paragraph{Dataset Construction.}
Our data split is designed to decouple rule acquisition from compositional reasoning.
%\yafu{add a case study of your data (training/testing), e.g., a table or a figure.}
\begin{itemize}
    \item \textbf{Training:} The training set is strictly limited to \textit{atomic} operations (1-hop reasoning). This forces the model to internalize the fundamental axioms of each group.
    \item \textbf{Testing:} The test set evaluates the model's ability to chain these axioms. It is stratified into four difficulty levels based on operation depth (2 to 5 ops), with 200 balanced samples per level across all domains.
\end{itemize}

Table~\ref{tab:experiment_overview} summarizes our dataset configuration. Further details for each task are provided in Appendix~\ref{app:domain_details}.

\paragraph{Implementation Details.}
We benchmark three model families: \textsc{Llama}, \textsc{Qwen}, and \textsc{Gemma}. Training employs {Group Relative Policy Optimization (GRPO, ~\citealp{Shao2024DeepSeekMathPT})} for 100 steps.  Performance is measured via exact match accuracy of the final boxed output. Further details are provided in Appendix~\ref{app:implementation_details}.

\section{Empirical Validation of the Theory}
 
\begin{figure*}[!ht]
    \centering
    \includegraphics[width=0.85\linewidth]{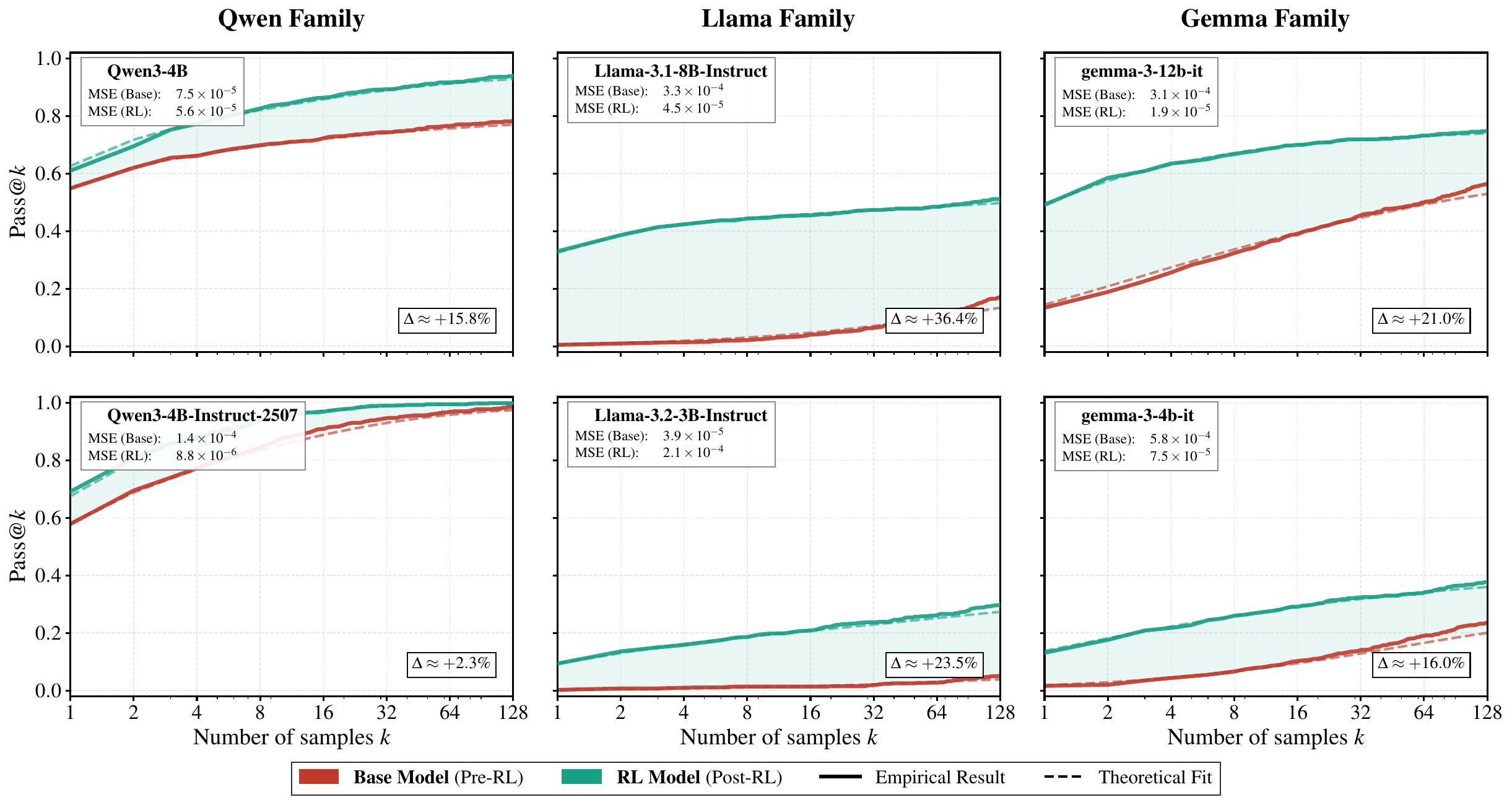}
    \caption{
        \textbf{Comparison of theoretical and empirical \textit{Pass@k} curves.} The figure illustrates that the theoretical and empirical curves align remarkably well, consistently yielding a Mean Squared Error (MSE) magnitude less than $10^{-4}$. Furthermore, the RL models demonstrate substantial performance improvements over the Base models across the sampling spectrum.
    }
    \label{fig:pass_k}
\end{figure*}

To validate the theory in Section~\ref{sec:theory_passk}, we evaluate \textsc{Llama}, \textsc{Qwen}, and \textsc{Gemma} models across several dimensions. Our experiments trace a causal chain from \textit{Pass@k Curve Dynamics} to the \textit{Multiplicative Barrier} of complex reasoning, showing how these drive \textit{Compositional Emergence} and, ultimately, \textit{Capability Erosion}.

\subsection{Validation of Theoretical \textit{Pass@k} Dynamics}
\label{sec:exp_passk}

First, we examine the fidelity of our theoretical model in capturing the empirical behavior of the \textit{Pass@k} metric. 
To construct the theoretical curves, we necessitate an estimate of the intrinsic correctness probability $P_\theta(q)$ for each problem. We approximate this parameter by computing the average correctness rate over 128 sampled responses for each instance (i.e., $P_\theta(q) \approx Avg@128(Q)$, using Eq.~\ref{eq:expect}).
The comparison between these theoretically projected curves and the actual empirical results is presented in Figure~\ref{fig:pass_k}.

%As illustrated in Figure~\ref{fig:pass_k}, the empirical trajectories strongly corroborate our theoretical visualization. The alignment is quantitatively precise, yielding a Mean Squared Error (MSE) consistently at the magnitude less than $10^{-4}$, which confirms the validity of the formulation described in Section~\ref{sec:theory_passk}.
%Furthermore, the visualization highlights a substantial improvement in the \textit{Pass@k} metric driven by RLVR. This performance leap implies the emergence of new capabilities, suggesting that the model has successfully transitioned previously intractable problems into the solvable regime.

As shown in Figure~\ref{fig:pass_k}, empirical results align with our theory with high precision, confirming the Section~\ref{sec:theory_passk} formulation. Significant improvements in \textit{Pass@k} suggest that RLVR enables the model to solve previously intractable problems, signaling the emergence of new capabilities.

\subsection{Transitions from the Null Regime}
\label{sec:exp_emergence}

Building on the probabilistic definitions established in Section~\ref{sec:capability_determination}, we investigate the mechanism of emergence. We rigorously isolate the subset of tasks initially classified within \textit{The Null Set} for the base models. To ensure evaluation robustness, we define the acquisition of capability strictly as the transition into \textit{The Feasible Set}. 

\begin{figure}[!ht]
    \centering
    \includegraphics[width=1\linewidth]{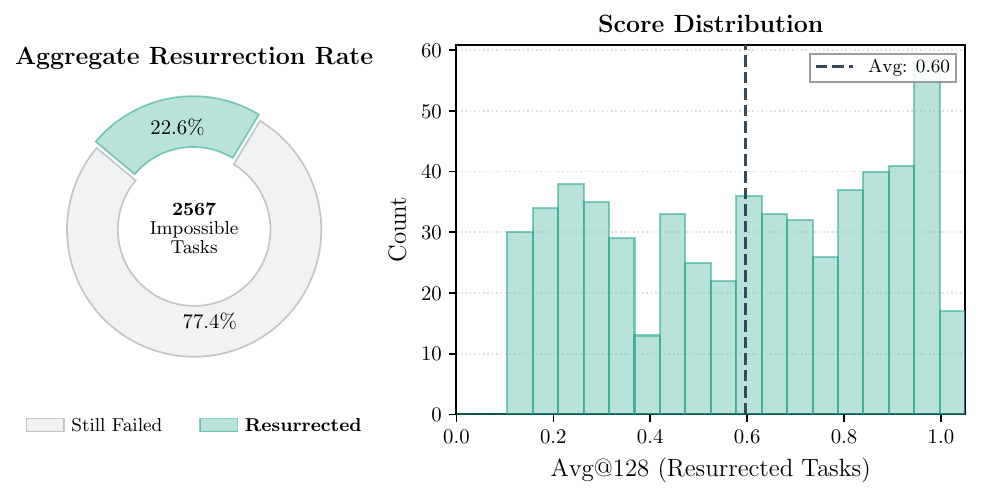}
    \caption{
     \textbf{Analysis of Capability Emergence.} 
      We analyze the subset of tasks initially classified as Null State ($\text{Avg@128} = 0$) in the Base model. 
      \textbf{Left:} The recovery rate, indicating the proportion of these impossible tasks that successfully transitioned to the Feasible State ($\text{Avg@128} \ge 0.125$) after RL training.
      \textbf{Right:} Post-emergence accuracy distribution. The significant skew toward high performance (mean $\approx 0.60$) indicates that emergence occurs as a sharp phase transition. 
    }   
    \label{fig:emergence-fig}
\end{figure}

As illustrated in Figure~\ref{fig:emergence-fig}, RLVR effectively unlocks 22.6\% of tasks from the Null Set, elevating the mean success rate of this activated subset from 0 to 0.60. This shift substantiates that emergence arises when optimized atomic probabilities surmount the multiplicative barrier, rendering previously intractable tasks reliably solvable.

\subsection{Validating the Multiplicative Barrier Hypothesis}

\label{sec:Multiplicative}

Section~\ref{sec:emergence_and_erosion} attributes the difficulty of complex tasks to the exponential decay of joint probabilities ($\prod P_\theta(s_j)$). We empirically validate this \textit{Multiplicative Barrier} empirically through two complementary analyses.

\paragraph{Existence of the Exponential Barrier.}

\begin{figure*}[!ht]
    \centering
    \includegraphics[width=0.85\linewidth]{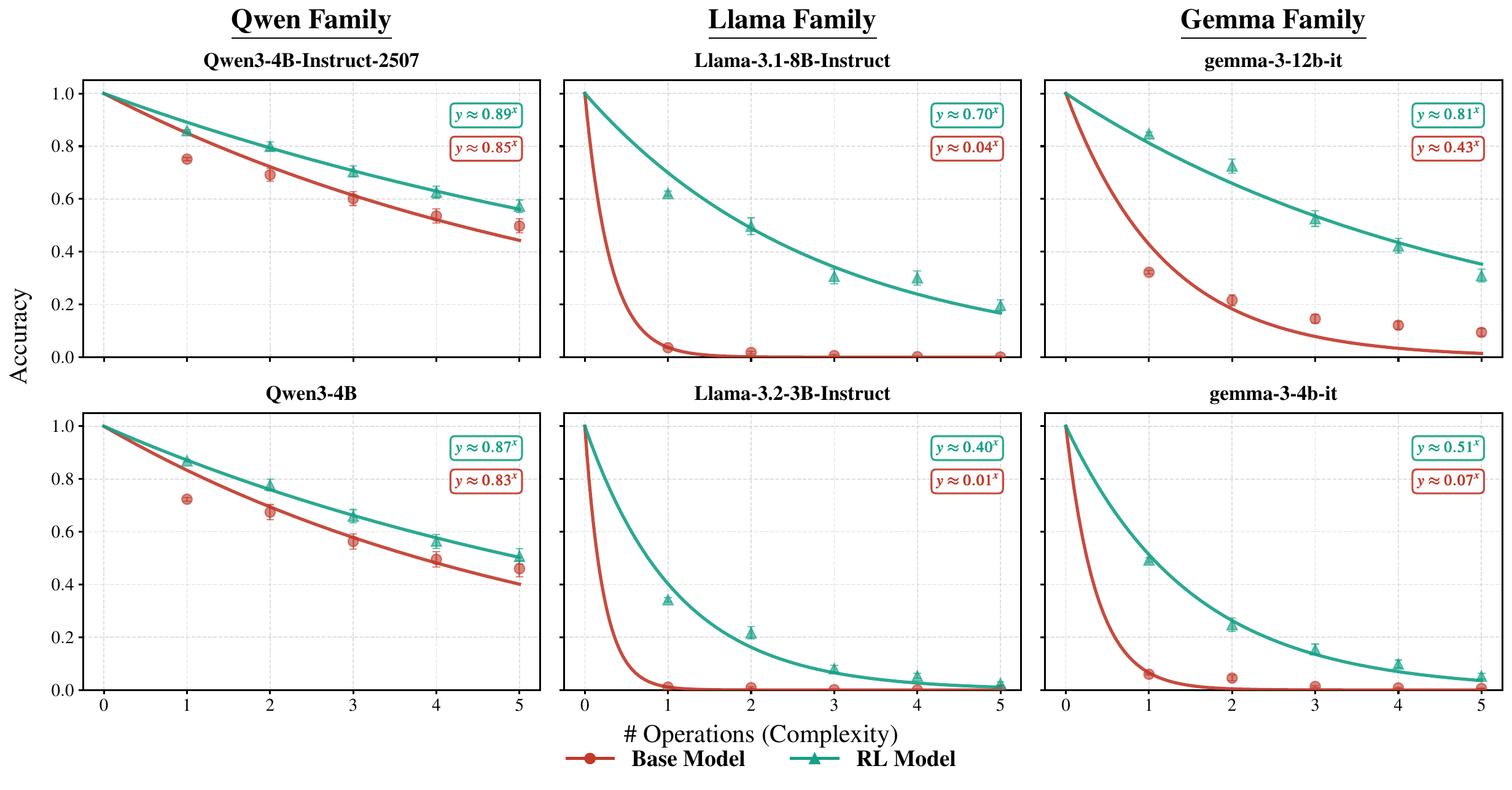}
    \caption{\textbf{Verification of Exponential Decay.} The figure validates the \textit{Multiplicative Barrier} hypothesis ($P \propto p^N$).}
    \label{fig:exp_decay}
\end{figure*}
\begin{figure*}[!ht]
    \centering
    \includegraphics[width=0.85\linewidth]{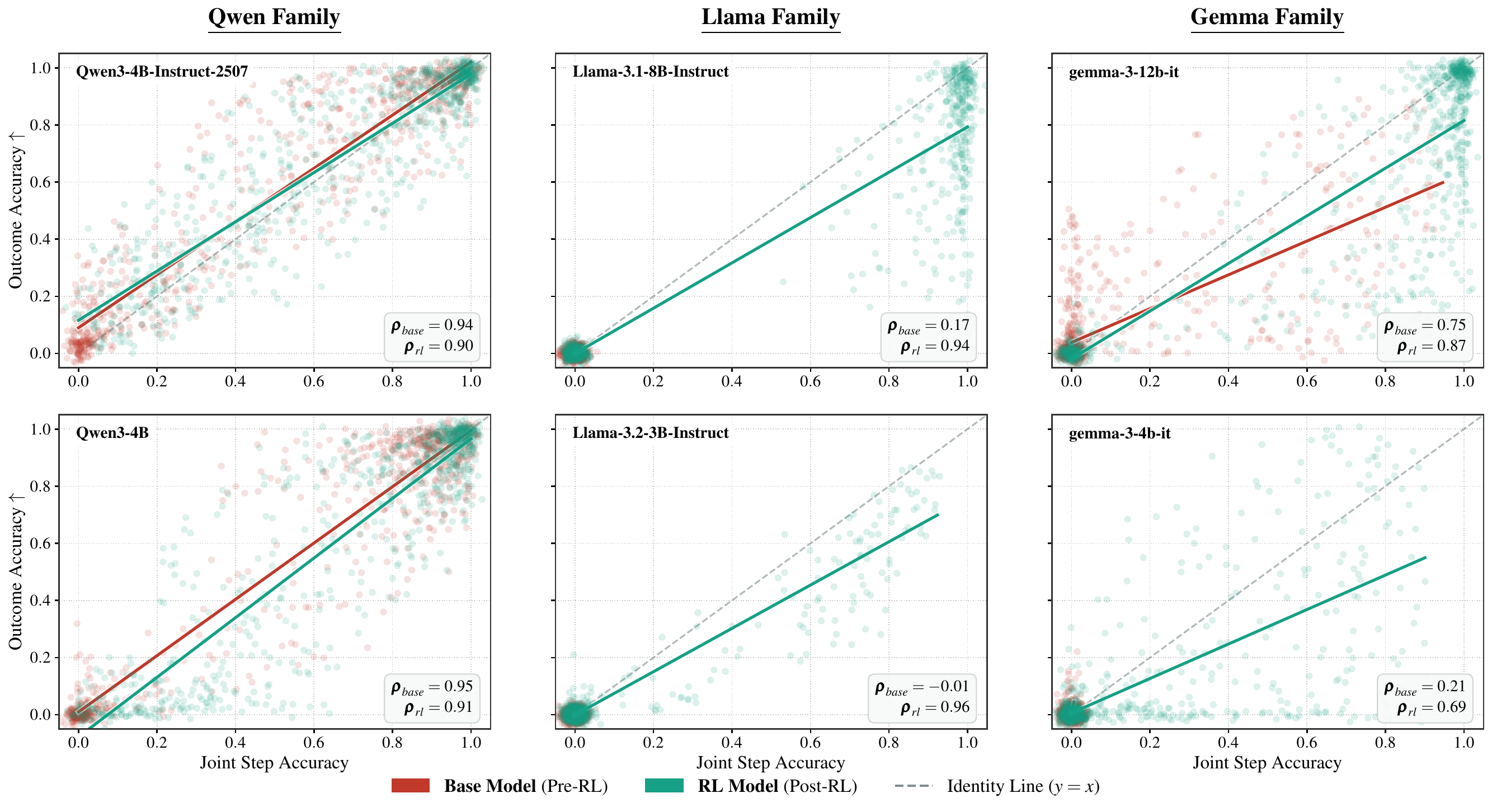}
    \caption{\textbf{Process vs. Outcome Correlation Analysis.} 
        Scatter plots correlating joint step accuracy ($\prod P_\theta(s_j)$) with Outcome Accuracy ($\prod P_\theta(q)$). Selected base models from the Llama and Gemma families cluster in the ``Null Regime" near the origin, indicating that a breakdown in the joint reasoning chain leads to task failure. In contrast, Qwen models and RL-tuned models exhibit a high correlation.
    }
    \label{fig:process_outcome}
\end{figure*}

Figure \ref{fig:exp_decay} delineates a distinct performance boundary, effectively acting as a ``soft wall" for model capabilities. We observe that accuracy undergoes a geometric decay relative to task complexity, closely following the function $y \approx p^x$. This trajectory provides empirical confirmation that reasoning errors accumulate multiplicatively: the joint probability of success is the product of individual atomic probabilities. Consequently, even with high atomic accuracy, the likelihood of a correct final outcome diminishes exponentially as the reasoning horizon extends, rendering highly complex tasks rapidly intractable.

\paragraph{Correlation as the Mechanism of Emergence.}
To uncover the mechanism behind the ``Zero-to-One" emergence phenomenon observed in Section~\ref{sec:exp_emergence}, Figure~\ref{fig:process_outcome} analyzes the correlation between joint step accuracy and final answer correctness. The data reveal that the emergence of complex capabilities is structurally bound by this correlation.

\begin{figure*}[!ht]
    \centering
    \includegraphics[width=0.85\linewidth]{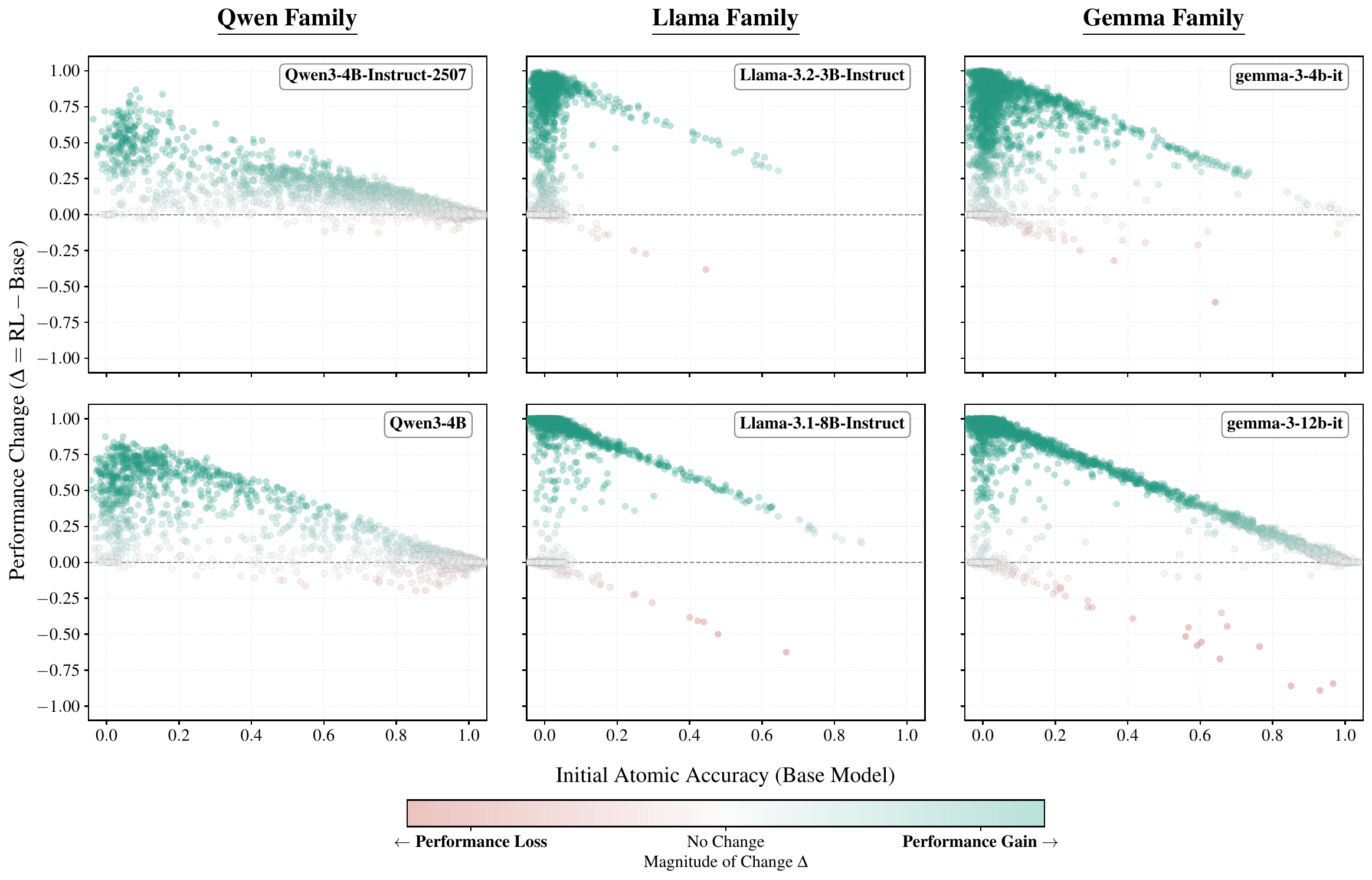} 
    \caption{\textbf{Analysis of Atomic Performance Shifts.} Visualization of performance change ($\Delta = \text{RL} - \text{Base}$) versus initial atomic accuracy. All model families achieve substantial performance gains across most atomic skills. However, Qwen demonstrates high robustness, achieving consistent gains with minimal regression in mastered skills. Conversely, Llama and Gemma exhibit a more extreme behavior. While they often achieve higher magnitude gains in weak areas, this often leads to a noticeable regression in previously acquired skills.}
    \label{fig:robin_hood}
\end{figure*}
Prior to RLVR, Gemma and Llama models exhibit a ``broken link" between process and outcome, clustering near the origin. Without sufficient atomic accuracy, the probabilistic chain collapses, preventing the capability from manifesting.

The transition to RLVR restores the linear coupling between atomic steps and final results. The high correlation coefficients ($\rho \in [0.69, 0.96]$) confirm that emergence is the direct mathematical consequence of pushing the joint process probability across the multiplicative barrier. 
Most critically, this indicates that complex capabilities introduced by RLVR are reducible to the mastery of constituent atomic tasks, suggesting a clear pathway for their acquisition.

\subsection{Capability Erosion under Atomic-Only RL}

\label{sec:atomic_erosion_exp}
\textcolor{black}{While RLVR fosters capability emergence, it can also induce performance erosion. As theorized in Section~\ref{theory:erosion}, its global optimization objective may sacrifice specialized skills to maximize aggregate rewards.}

We examine the optimization dynamics in Figure~\ref{fig:robin_hood} by plotting the performance change ($\Delta$) against the initial accuracy for each atomic skill.
As the initial mastery of a skill increases, the difficulty of further improvement rises, and the risk of regression emerges. However, the severity of this trade-off varies significantly.

\textbf{Llama \& Gemma (High Severity).}
These families show an extreme exchange. While they achieve substantial gains on initially weak skills, they pay a visible price in stability. The scatter plots reveal distinct clusters of negative values, indicating that the acquisition of new capabilities often comes at the expense of eroding previously mastered skills.

\textbf{Qwen (Mitigated Severity).}
Qwen manages this trade-off more effectively. While it also faces diminishing returns on high-accuracy skills, it avoids the catastrophic drops seen in others. This stability reflects that Qwen samples responses with stronger logical consistency. This internal consistency acts as a buffer, allowing Qwen to navigate the RLVR process and improve weak areas without largely sacrificing the integrity of its existing skills.

%\section{Discussion}

% A metric， given a model without changing model inherent parameters, can predict downstream RL performance.

\section{Conclusion}

In this paper, we explain how RLVR enables the emergence of new capabilities. 
We reveal that instance-level correctness is the true driver of \textit{Pass@k} performance. We identify the \textit{Multiplicative Barrier} in complex reasoning, where the probability of success decays exponentially as reasoning chains lengthen. Then, we demonstrate that models overcome this by optimizing their existing capabilities. This optimization allows the model to discover valid reasoning paths that were previously inaccessible. 
Overall, our work contributes a novel explanation to the current discussion on emergent abilities in RLVR, suggesting that the iterative optimization of solvable problems enables models to develop the capabilities to tackle previously unsolvable scenarios.

%\section*{Accessibility}

%Authors are kindly asked to make their submissions as accessible as possible
%for everyone including people with disabilities and sensory or neurological
%differences. Tips of how to achieve this and what to pay attention to will be
%provided on the conference website \url{http://icml.cc/}.

% \section*{Software and Data}

% If a paper is accepted, we strongly encourage the publication of software and
% data with the camera-ready version of the paper whenever appropriate. This can
% be done by including a URL in the camera-ready copy. However, \textbf{do not}
% include URLs that reveal your institution or identity in your submission for
% review. Instead, provide an anonymous URL or upload the material as
% ``Supplementary Material'' into the OpenReview reviewing system. Note that
% reviewers are not required to look at this material when writing their review.

% \section*{Acknowledgements}

\section*{Impact Statement}
This paper advances the understanding of how reinforcement learning facilitates complex reasoning in LLMs. By theoretically modeling the "Multiplicative Barrier" and clarifying the true drivers of performance beyond \textit{Pass@k}, we provide actionable insights for improving model efficiency and correctness. These findings are instrumental for the research community to develop more effective training paradigms, ultimately leading to AI assistants that are capable of solving intricate problems with higher verifiable accuracy, benefiting fields ranging from automated coding to formal logic.

\newpage

\bibliography{example_paper}

@article{ElKishky2024OpenAIOS,
  title={OpenAI o1 System Card},
  author={OpenAI},
  journal={ArXiv},
  year={2024},
}

@article{DeepSeekAI2025DeepSeekR1IR,
  title={DeepSeek-R1: Incentivizing Reasoning Capability in LLMs via Reinforcement Learning},
  author={DeepSeek-AI and Daya Guo and Dejian Yang and Haowei Zhang and Jun-Mei Song and Ruoyu Zhang and Runxin Xu and Qihao Zhu and Shirong Ma and Peiyi Wang and Xiaoling Bi and Xiaokang Zhang and Xingkai Yu and Yu Wu and Z. F. Wu and Zhibin Gou and Zhihong Shao and Zhuoshu Li and Ziyi Gao and Aixin Liu and Bing Xue and Bing-Li Wang and Bochao Wu and Bei Feng and Chengda Lu and Chenggang Zhao and Chengqi Deng and Chenyu Zhang and Chong Ruan and Damai Dai and Deli Chen and Dong-Li Ji and Erhang Li and Fangyun Lin and Fucong Dai and Fuli Luo and Guangbo Hao and Guanting Chen and Guowei Li and H. Zhang and Han Bao and Hanwei Xu and Haocheng Wang and Honghui Ding and Huajian Xin and Huazuo Gao and Hui Qu and Hui Li and Jianzhong Guo and Jiashi Li and Jiawei Wang and Jingchang Chen and Jingyang Yuan and Junjie Qiu and Junlong Li and Jiong Cai and Jiaqi Ni and Jian Liang and Jin Chen and Kai Dong and Kai Hu and Kaige Gao and Kang Guan and Kexin Huang and Kuai Yu and Lean Wang and Lecong Zhang and Liang Zhao and Litong Wang and Liyue Zhang and Lei Xu and Leyi Xia and Mingchuan Zhang and Minghua Zhang and M. Tang and Meng Li and Miaojun Wang and Mingming Li and Ning Tian and Panpan Huang and Peng Zhang and Qiancheng Wang and Qinyu Chen and Qiushi Du and Ruiqi Ge and Ruisong Zhang and Ruizhe Pan and Runji Wang and R. J. Chen and Ruiqi Jin and Ruyi Chen and Shanghao Lu and Shangyan Zhou and Shanhuang Chen and Shengfeng Ye and Shiyu Wang and Shuiping Yu and Shunfeng Zhou and Shuting Pan and S. S. Li and Shuang Zhou and Shao-Kang Wu and Tao Yun and Tian Pei and Tianyu Sun and T. Wang and Wangding Zeng and Wanjia Zhao and Wen Liu and Wenfeng Liang and Wenjun Gao and Wen-Xia Yu and Wentao Zhang and Wangding Xiao and Wei An and Xiaodong Liu and Xiaohan Wang and Xi-aokang Chen and Xiaotao Nie and Xin Cheng and Xin Liu and Xin Xie and Xingchao Liu and Xinyu Yang and Xinyuan Li and Xuecheng Su and Xuheng Lin and X. Q. Li and Xiangyu Jin and Xi-Cheng Shen and Xiaosha Chen and Xiaowen Sun and Xiaoxiang Wang and Xinnan Song and Xinyi Zhou and Xianzu Wang and Xinxia Shan and Y. K. Li and Y. Q. Wang and Y. X. Wei and Yang Zhang and Yanhong Xu and Yao Li and Yao Zhao and Yaofeng Sun and Yaohui Wang and Yi Yu and Yichao Zhang and Yifan Shi and Yi Xiong and Ying He and Yishi Piao and Yisong Wang and Yixuan Tan and Yiyang Ma and Yiyuan Liu and Yongqiang Guo and Yuan Ou and Yuduan Wang and Yue Gong and Yu-Jing Zou and Yujia He and Yunfan Xiong and Yu-Wei Luo and Yu-mei You and Yuxuan Liu and Yuyang Zhou and Y. X. Zhu and Yanping Huang and Yao Li and Yi Zheng and Yuchen Zhu and Yunxiang Ma and Ying Tang and Yukun Zha and Yuting Yan and Zehui Ren and Zehui Ren and Zhangli Sha and Zhe Fu and Zhean Xu and Zhenda Xie and Zhen-guo Zhang and Zhewen Hao and Zhicheng Ma and Zhigang Yan and Zhiyu Wu and Zihui Gu and Zijia Zhu and Zijun Liu and Zi-An Li and Ziwei Xie and Ziyang Song and Zizheng Pan and Zhen Huang and Zhipeng Xu and Zhongyu Zhang and Zhen Zhang},
  journal={Nature},
  year={2025},
  volume={645},
}

@article{GDM2025Gemini2P,
  title={Gemini 2.5: Pushing the Frontier with Advanced Reasoning, Multimodality, Long Context, and Next Generation Agentic Capabilities},
  author={Google DeepMind},
  journal={ArXiv},
  year={2025},
  volume={abs/2507.06261},
}

@misc{wei2023chainofthoughtpromptingelicitsreasoning,
      title={Chain-of-Thought Prompting Elicits Reasoning in Large Language Models}, 
      author={Jason Wei and Xuezhi Wang and Dale Schuurmans and Maarten Bosma and Brian Ichter and Fei Xia and Ed Chi and Quoc Le and Denny Zhou},
      year={2023},
      eprint={2201.11903},
      archivePrefix={arXiv},
      primaryClass={cs.CL},
      url={https://arxiv.org/abs/2201.11903}, 
}

@article{Gandhi2025CognitiveBT,
  title={Cognitive Behaviors that Enable Self-Improving Reasoners, or, Four Habits of Highly Effective STaRs},
  author={Kanishk Gandhi and Ayush Chakravarthy and Anikait Singh and nathan lile and Noah D. Goodman},
  journal={COLM},
  year={2025},
}

@article{Yue2025DoesRL,
  title={Does Reinforcement Learning Really Incentivize Reasoning Capacity in LLMs Beyond the Base Model?},
  author={Yang Yue and Zhiqi Chen and Rui Lu and Andrew Zhao and Zhaokai Wang and Shiji Song and Gao Huang},
  journal={ArXiv},
  year={2025},
}

@article{DBLP:journals/corr/abs-1910-07113,
  author       = {OpenAI and
                  Ilge Akkaya and
                  Marcin Andrychowicz and
                  Maciek Chociej and
                  Mateusz Litwin and
                  Bob McGrew and
                  Arthur Petron and
                  Alex Paino and
                  Matthias Plappert and
                  Glenn Powell and
                  Raphael Ribas and
                  Jonas Schneider and
                  Nikolas Tezak and
                  Jerry Tworek and
                  Peter Welinder and
                  Lilian Weng and
                  Qiming Yuan and
                  Wojciech Zaremba and
                  Lei Zhang},
  title        = {Solving Rubik's Cube with a Robot Hand},
  journal      = {CoRR},
  volume       = {abs/1910.07113},
  year         = {2019},
  url          = {http://arxiv.org/abs/1910.07113},
  eprinttype    = {arXiv},
  eprint       = {1910.07113},
  bibsource    = {dblp computer science bibliography, https://dblp.org}
}

@article{Zhao2025EchoChamber,
  author       = {Rosie Zhao and
                  Alexandru Meterez and
                  Sham M. Kakade and
                  Cengiz Pehlevan and
                  Samy Jelassi and
                  Eran Malach},
  title        = {Echo Chamber: {RL} Post-training Amplifies Behaviors Learned in Pretraining},
  journal      = {CoRR},
  volume       = {abs/2504.07912},
  year         = {2025},
  url          = {https://doi.org/10.48550/arXiv.2504.07912},
  doi          = {10.48550/ARXIV.2504.07912},
  eprinttype    = {arXiv},
  eprint       = {2504.07912},
  bibsource    = {dblp computer science bibliography, https://dblp.org}
}

@article{Liu2025DrGRPO,
  author       = {Zichen Liu and
                  Changyu Chen and
                  Wenjun Li and
                  Penghui Qi and
                  Tianyu Pang and
                  Chao Du and
                  Wee Sun Lee and
                  Min Lin},
  title        = {Understanding R1-Zero-Like Training: {A} Critical Perspective},
  journal      = {COLM},
  year         = {2025},
}

@article{Shao2024DeepSeekMathPT,
  title={DeepSeekMath: Pushing the Limits of Mathematical Reasoning in Open Language Models},
  author={Zhihong Shao and Peiyi Wang and Qihao Zhu and Runxin Xu and Jun-Mei Song and Mingchuan Zhang and Y. K. Li and Yu Wu and Daya Guo},
  journal={ArXiv},
  year={2024},
  volume={abs/2402.03300},
}

@article{he2025rewarding,
  title={Rewarding the Unlikely: Lifting GRPO Beyond Distribution Sharpening},
  author={He, Andre and Fried, Daniel and Welleck, Sean},
  journal={arXiv preprint arXiv:2506.02355},
  year={2025}
}

@article{wen2025reinforcement,
  title={Reinforcement Learning with Verifiable Rewards Implicitly Incentivizes Correct Reasoning in Base LLMs},
  author={Wen, Xumeng and Liu, Zihan and Zheng, Shun and Xu, Zhijian and Ye, Shengyu and Wu, Zhirong and Liang, Xiao and Wang, Yang and Li, Junjie and Miao, Ziming and others},
  journal={arXiv preprint arXiv:2506.14245},
  year={2025}
}

@article{zhu2025surprising,
  title={The surprising effectiveness of negative reinforcement in LLM reasoning},
  author={Zhu, Xinyu and Xia, Mengzhou and Wei, Zhepei and Chen, Wei-Lin and Chen, Danqi and Meng, Yu},
  journal={arXiv preprint arXiv:2506.01347},
  year={2025}
}

@misc{yuan2025fxgxfgxllms,
      title={From $f(x)$ and $g(x)$ to $f(g(x))$: LLMs Learn New Skills in RL by Composing Old Ones}, 
      author={Lifan Yuan and Weize Chen and Yuchen Zhang and Ganqu Cui and Hanbin Wang and Ziming You and Ning Ding and Zhiyuan Liu and Maosong Sun and Hao Peng},
      year={2025},
      eprint={2509.25123},
      archivePrefix={arXiv},
      primaryClass={cs.AI},
      url={https://arxiv.org/abs/2509.25123}, 
}

@misc{cheng2025atomiccompositereinforcementlearning,
      title={From Atomic to Composite: Reinforcement Learning Enables Generalization in Complementary Reasoning}, 
      author={Sitao Cheng and Xunjian Yin and Ruiwen Zhou and Yuxuan Li and Xinyi Wang and Liangming Pan and William Yang Wang and Victor Zhong},
      year={2025},
      eprint={2512.01970},
      archivePrefix={arXiv},
      primaryClass={cs.AI},
      url={https://arxiv.org/abs/2512.01970}, 
}

@misc{dziri2023faithfatelimitstransformers,
      title={Faith and Fate: Limits of Transformers on Compositionality}, 
      author={Nouha Dziri and Ximing Lu and Melanie Sclar and Xiang Lorraine Li and Liwei Jiang and Bill Yuchen Lin and Peter West and Chandra Bhagavatula and Ronan Le Bras and Jena D. Hwang and Soumya Sanyal and Sean Welleck and Xiang Ren and Allyson Ettinger and Zaid Harchaoui and Yejin Choi},
      year={2023},
      eprint={2305.18654},
      archivePrefix={arXiv},
      primaryClass={cs.CL},
      url={https://arxiv.org/abs/2305.18654}, 
}

@misc{zhou2023limaalignment,
      title={LIMA: Less Is More for Alignment}, 
      author={Chunting Zhou and Pengfei Liu and Puxin Xu and Srini Iyer and Jiao Sun and Yuning Mao and Xuezhe Ma and Avia Efrat and Ping Yu and Lili Yu and Susan Zhang and Gargi Ghosh and Mike Lewis and Luke Zettlemoyer and Omer Levy},
      year={2023},
      eprint={2305.11206},
      archivePrefix={arXiv},
      primaryClass={cs.CL},
      url={https://arxiv.org/abs/2305.11206}, 
}

@misc{cui2025entropymechanismreinforcementlearning,
      title={The Entropy Mechanism of Reinforcement Learning for Reasoning Language Models}, 
      author={Ganqu Cui and Yuchen Zhang and Jiacheng Chen and Lifan Yuan and Zhi Wang and Yuxin Zuo and Haozhan Li and Yuchen Fan and Huayu Chen and Weize Chen and Zhiyuan Liu and Hao Peng and Lei Bai and Wanli Ouyang and Yu Cheng and Bowen Zhou and Ning Ding},
      year={2025},
      eprint={2505.22617},
      archivePrefix={arXiv},
      primaryClass={cs.LG},
      url={https://arxiv.org/abs/2505.22617}, 
}

@article{wang2025beyond,
  title={Beyond the 80/20 rule: High-entropy minority tokens drive effective reinforcement learning for llm reasoning},
  author={Wang, Shenzhi and Yu, Le and Gao, Chang and Zheng, Chujie and Liu, Shixuan and Lu, Rui and Dang, Kai and Chen, Xionghui and Yang, Jianxin and Zhang, Zhenru and others},
  journal={arXiv preprint arXiv:2506.01939},
  year={2025}
}

@misc{zhan2025exgrpolearningreasonexperience,
      title={ExGRPO: Learning to Reason from Experience}, 
      author={Runzhe Zhan and Yafu Li and Zhi Wang and Xiaoye Qu and Dongrui Liu and Jing Shao and Derek F. Wong and Yu Cheng},
      year={2025},
      eprint={2510.02245},
      archivePrefix={arXiv},
      primaryClass={cs.LG},
      url={https://arxiv.org/abs/2510.02245}, 
}

@misc{park2025doesrlposttraininginduce,
      title={How Does RL Post-training Induce Skill Composition? A Case Study on Countdown}, 
      author={Simon Park and Simran Kaur and Sanjeev Arora},
      year={2025},
      eprint={2512.01775},
      archivePrefix={arXiv},
      primaryClass={cs.LG},
      url={https://arxiv.org/abs/2512.01775}, 
}

@misc{yu2025dapoopensourcellmreinforcement,
      title={DAPO: An Open-Source LLM Reinforcement Learning System at Scale}, 
      author={Qiying Yu and Zheng Zhang and Ruofei Zhu and Yufeng Yuan and Xiaochen Zuo and Yu Yue and Weinan Dai and Tiantian Fan and Gaohong Liu and Lingjun Liu and Xin Liu and Haibin Lin and Zhiqi Lin and Bole Ma and Guangming Sheng and Yuxuan Tong and Chi Zhang and Mofan Zhang and Wang Zhang and Hang Zhu and Jinhua Zhu and Jiaze Chen and Jiangjie Chen and Chengyi Wang and Hongli Yu and Yuxuan Song and Xiangpeng Wei and Hao Zhou and Jingjing Liu and Wei-Ying Ma and Ya-Qin Zhang and Lin Yan and Mu Qiao and Yonghui Wu and Mingxuan Wang},
      year={2025},
      eprint={2503.14476},
      archivePrefix={arXiv},
      primaryClass={cs.LG},
      url={https://arxiv.org/abs/2503.14476}, 
}

@misc{zheng2025groupsequencepolicyoptimization,
      title={Group Sequence Policy Optimization}, 
      author={Chujie Zheng and Shixuan Liu and Mingze Li and Xiong-Hui Chen and Bowen Yu and Chang Gao and Kai Dang and Yuqiong Liu and Rui Men and An Yang and Jingren Zhou and Junyang Lin},
      year={2025},
      eprint={2507.18071},
      archivePrefix={arXiv},
      primaryClass={cs.LG},
      url={https://arxiv.org/abs/2507.18071}, 
}

@misc{yan2025learningreasonoffpolicyguidance,
      title={Learning to Reason under Off-Policy Guidance}, 
      author={Jianhao Yan and Yafu Li and Zican Hu and Zhi Wang and Ganqu Cui and Xiaoye Qu and Yu Cheng and Yue Zhang},
      year={2025},
      eprint={2504.14945},
      archivePrefix={arXiv},
      primaryClass={cs.LG},
      url={https://arxiv.org/abs/2504.14945}, 
}

@misc{jiang2025rethinkingentropyregularizationlarge,
      title={Rethinking Entropy Regularization in Large Reasoning Models}, 
      author={Yuxian Jiang and Yafu Li and Guanxu Chen and Dongrui Liu and Yu Cheng and Jing Shao},
      year={2025},
      eprint={2509.25133},
      archivePrefix={arXiv},
      primaryClass={cs.LG},
      url={https://arxiv.org/abs/2509.25133}, 
}
\bibliographystyle{icml2025}

\newpage
%%%%%%%%%%%%%%%%%%%%%%%%%%%%%%%%%%%%%%%%%%%%%%%%%%%%%%%%%%%%%%%%%%%%%%%%%%%%%%%
%%%%%%%%%%%%%%%%%%%%%%%%%%%%%%%%%%%%%%%%%%%%%%%%%%%%%%%%%%%%%%%%%%%%%%%%%%%%%%%
% APPENDIX
%%%%%%%%%%%%%%%%%%%%%%%%%%%%%%%%%%%%%%%%%%%%%%%%%%%%%%%%%%%%%%%%%%%%%%%%%%%%%%%
%%%%%%%%%%%%%%%%%%%%%%%%%%%%%%%%%%%%%%%%%%%%%%%%%%%%%%%%%%%%%%%%%%%%%%%%%%%%%%%

\appendix

\onecolumn
\section{Algebrarium Framework}
\label{app:algebrarium}

To rigorously evaluate the symbolic reasoning capabilities of Large Language Models (LLMs) while mitigating the risks of data contamination, we developed \textbf{Algebrarium}. Unlike standard benchmarks that rely on static datasets, which are susceptible to memorization during pre-training, Algebrarium serves as a framework for the \textbf{procedural generation of synthetic algebraic tasks}. By constructing data dynamically based on arbitrary, user-defined mathematical structures, it establishes a hermetic evaluation environment where models must rely on on-the-fly reasoning rather than retrieval.

The core philosophy of Algebrarium lies in the decoupling of \textit{abstract mathematical definition} from \textit{concrete task generation}. This approach facilitates the creation of infinite variations of reasoning problems derived from a single logical core. The framework adheres to three fundamental principles:

\paragraph{1. Arbitrary Algebraic Structures}
Algebrarium treats algebra as a formal system defined by a tuple $(S, \Sigma, \mathcal{A})$, where $S$ denotes a carrier set (finite or infinite), $\Sigma$ represents a signature of operations with defined arities and precedence, and $\mathcal{A}$ constitutes the axioms or computational rules governing interaction. Crucially, these structures are defined declaratively. This design allows the framework to generate tasks for standard structures, such as modular arithmetic or non-abelian groups, as well as entirely novel and custom-defined logical systems that the model has never encountered. Consequently, this enforces genuine zero-shot reasoning.

\paragraph{2. Procedural Task Construction}
Rather than retrieving pre-existing questions, the framework constructs reasoning tasks algorithmically. A generation engine instantiates specific algebraic instances and produces symbolic artifacts, such as expression trees or equations, to ensure that every sample is unique. This pipeline includes mechanisms for:
\begin{itemize}
    \item \textbf{Controllable Complexity:} Parameters including recursion depth, expression length, and the cardinality of the underlying set can be adjusted to create a smooth difficulty gradient.
    \item \textbf{Diverse Task Types:} The engine supports various symbolic tasks, ranging from forward evaluation, such as reducing an expression $a \circ b \circ c$ to a single element, to inverse reasoning, which involves solving equations like $a \circ x = b$ or systems of group equations.
\end{itemize}

\paragraph{3. Verifiable Ground Truth}
Since the data is generated from formal logic rules, the ``ground truth'' is deterministic and mathematically verifiable. This characteristic allows for exact-match evaluation of the model's outputs and enables the automatic generation of step-by-step Chain-of-Thought (CoT) traces for fine-tuning or interpretability studies.

\section{Implementation Details and Training Dynamics}
\label{app:implementation_details}

\subsection{Base Models and Infrastructure}
We evaluate a comprehensive suite of base models to ensure the generalizability of our findings. Our experiments cover three major model families:
\begin{itemize}
    \item \textbf{Qwen3:} Qwen3-4B,  and Qwen3-4B-Instruct.
    \item \textbf{Llama-3.1:} Llama-3.1-8B-Instruct and Llama-3.1-3B-Instruct.
    \item \textbf{Gemma-3:} gemma-3-4b-it and gemma-3-12b-it.
\end{itemize}
All models are trained using the \texttt{verl} framework. We utilize \texttt{vLLM} for efficient rollout generation, employing Tensor Parallelism ($TP=2$) to handle the memory requirements of long-context reasoning up to 8,192 tokens.

\subsection{Hyperparameter Configuration}
For all experiments, we employ Group Relative Policy Optimization (GRPO) using the AdamW optimizer with a learning rate of $1 \times 10^{-6}$ over 100 optimization steps. The training is conducted with a global batch size of 128, a mini-batch size of 8, and a group size ($G$) of 8, with gradient accumulation enabled. We set the maximum prompt and response lengths to 1,024 and 8,192 tokens, respectively. Regarding the RL objective, we monitor the KL divergence with a penalty coefficient $\beta=0.001$, though the KL loss is not included in the optimization objective. Additionally, advantage normalization is disabled during training.

\subsection{Training Dynamics}
\label{app:training_dynamics}

\begin{figure*}[h]
    \centering
    \includegraphics[width=\textwidth]{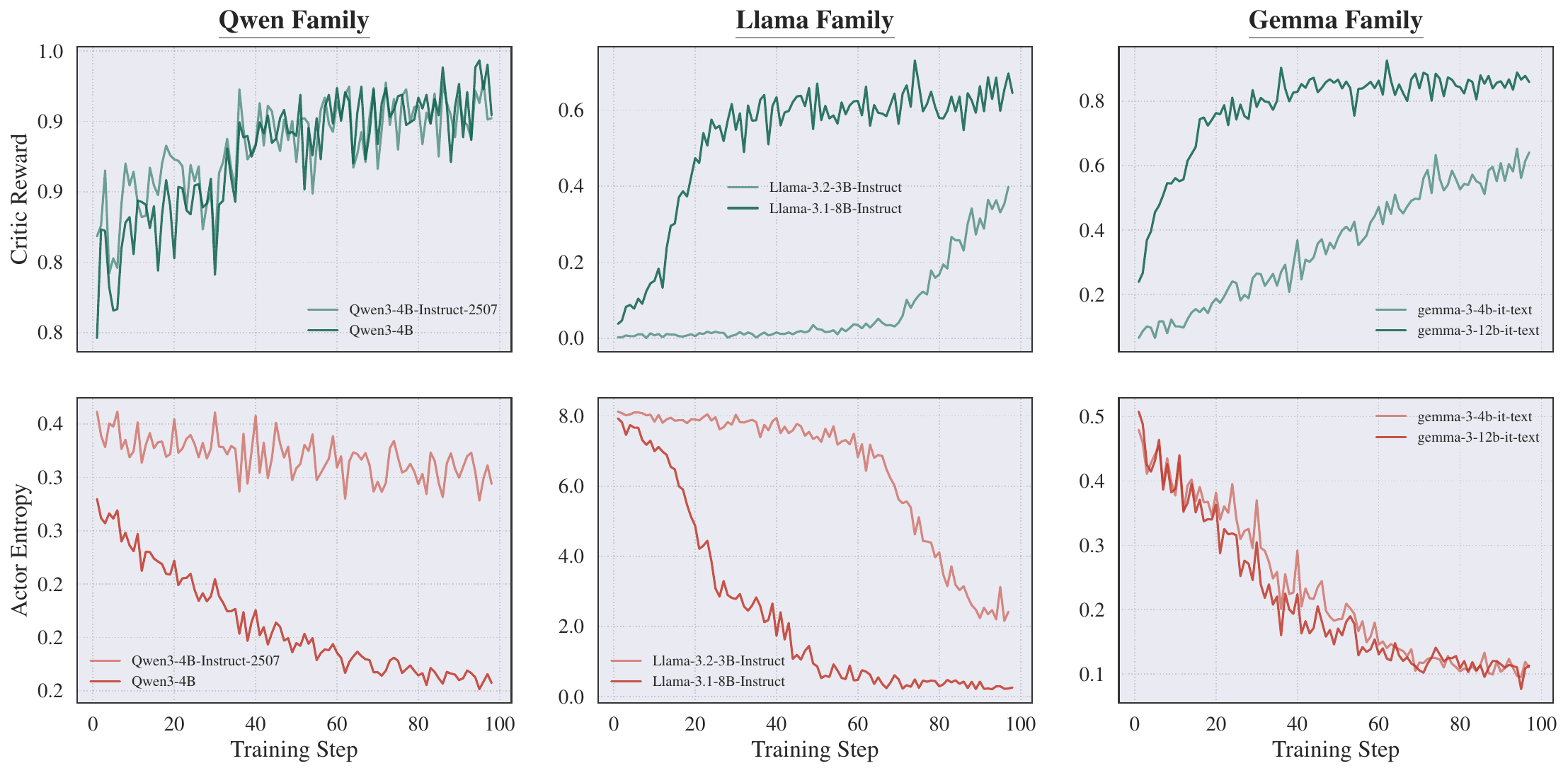}
    \caption{\textbf{Training Dynamics across Model Families.} The top row displays the evolution of Critic Reward, while the bottom row illustrates the Actor Entropy over 100 training steps. The results demonstrate consistent convergence behaviors across Qwen, Llama, and Gemma families.}
    \label{fig:training_curves}
\end{figure*}

We analyze the training stability and performance progression throughout the 100 optimization steps. As illustrated in Figure~\ref{fig:training_curves}, we monitor two key metrics across the Qwen, Llama, and Gemma model families:

\begin{itemize}
    \item \textbf{Critic Reward (Top Row):} The trajectory of the mean reward (Pass@1) starts near zero and exhibits a sharp increase before converging, indicating the models are effectively learning the target objective.
    \item \textbf{Actor Entropy (Bottom Row):} The policy entropy decreases rapidly during the initial phase and stabilizes as training progresses, reflecting the increasing confidence of the policy distribution.
\end{itemize}

\section{Domain Specifications and Operational Semantics}
\label{app:domain_details}

In this section, we provide the detailed formal specifications for the four algebraic systems employed in our experiments. Each system is constructed via the \textbf{Algebrarium} framework using distinct generative logic to cover different quadrants of group theory properties (Abelian vs. Non-Abelian, Finite vs. Infinite).

\subsection{Encrypted History Navigation (Infinite Abelian Group)}
This domain models 1-dimensional navigation on an integer line, obfuscated by a substitution cipher. It tests the model's ability to perform arithmetic operations ($\mathbb{Z}$) under symbolic encryption.

\paragraph{Representation}
\begin{itemize}
    \item \textbf{Base System:} Base-7.
    \item \textbf{Cipher Alphabet:} A bijective mapping $\phi: \{0, \dots, 6\} \to \{a, \dots, g\}$ where $0 \mapsto a, 1 \mapsto b, \dots, 6 \mapsto g$.
    \item \textbf{Structure:} Elements are strings in the format $\texttt{DIR(val)}$, where $\texttt{DIR} \in \{\texttt{FWD}, \texttt{BACK}\}$ and $\texttt{val}$ is a string of cipher characters representing a Base-7 magnitude.
\end{itemize}

\paragraph{Operational Logic}
Given two navigation steps $S_1$ and $S_2$:
\begin{enumerate}
    \item \textbf{Decryption:} The cipher string is decoded into a Base-10 integer $N = \sum_{i=0}^{L-1} \phi^{-1}(c_i) \cdot 7^{(L-1-i)}$.
    \item \textbf{Signed Interpretation:} $\texttt{FWD}$ implies $+N$, while $\texttt{BACK}$ implies $-N$.
    \item \textbf{Arithmetic Combination:} We compute $R = N_1 + N_2$.
    \item \textbf{Re-encryption:} The result $R$ is encoded back into the target format:
    \begin{itemize}
        \item If $R > 0$, the direction is $\texttt{FWD}$.
        \item If $R < 0$, the direction is $\texttt{BACK}$.
        \item The absolute value $|R|$ is converted to Base-7 and mapped via $\phi$.
    \end{itemize}
\end{enumerate}

\paragraph{Example}
Computing $\texttt{FWD(ad)} + \texttt{BACK(ef)}$:
\begin{itemize}
    \item $\texttt{ad}_7 \rightarrow 0\cdot7 + 3 = 3$. Direction FWD $\rightarrow +3$.
    \item $\texttt{ef}_7 \rightarrow 4\cdot7 + 5 = 33$. Direction BACK $\rightarrow -33$.
    \item Sum: $3 + (-33) = -30$.
    \item Result: Magnitude $30 = 42_7 \rightarrow \texttt{ec}$. Sign Negative $\rightarrow \texttt{BACK}$. Final: $\texttt{BACK(ec)}$.
\end{itemize}

\subsection{Enigma 3-Rotor System (Finite Abelian Group)}
This domain simulates the state of a simplified Enigma machine with three independent rotors. Mathematically, it represents the direct product group $\mathbb{Z}_{26} \times \mathbb{Z}_{26} \times \mathbb{Z}_{26}$.

\paragraph{Representation}
\begin{itemize}
    \item \textbf{State:} A comma-separated string $R_1,R_2,R_3$ where each $R_i \in \{A, \dots, Z\}$.
    \item \textbf{Mapping:} Letters map to offsets $0 \dots 25$ (A=0, B=1, ... Z=25).
\end{itemize}

\paragraph{Operational Logic}
The binary operation combines two states via component-wise addition modulo 26. Unlike a physical Enigma machine, there is no "carry" or stepping mechanism between rotors in this simplified algebraic model.
\begin{equation}
    (r_{1}, r_{2}, r_{3}) \oplus (s_{1}, s_{2}, s_{3}) = ((r_1+s_1)\text{mod}_{26}, (r_2+s_2)\text{mod}_{26}, (r_3+s_3)\text{mod}_{26})
\end{equation}
The inverse operation corresponds to negation modulo 26: $\text{inv}(r) = (26 - r) \pmod{26}$.

\paragraph{Example}
Computing $\texttt{A,C,Z} \oplus \texttt{B,B,C}$:
\begin{itemize}
    \item Rotor 1: $A(0) + B(1) = 1 \rightarrow \texttt{B}$
    \item Rotor 2: $C(2) + B(1) = 3 \rightarrow \texttt{D}$
    \item Rotor 3: $Z(25) + C(2) = 27 \equiv 1 \pmod{26} \rightarrow \texttt{B}$
    \item Result: $\texttt{B,D,B}$
\end{itemize}

\subsection{Knitting Instruction System (Infinite Non-Abelian Group)}
This domain models a Free Group generated by two elements. It represents sequences of reversible actions where order matters and adjacent inverses cancel.

\paragraph{Representation}
\begin{itemize}
    \item \textbf{Alphabet:} $\Sigma = \{k, p, K, P\}$.
    \item \textbf{Semantics:} $k$ (knit), $p$ (purl). Capital letters represent inverses: $K = k^{-1}$ (un-knit), $P = p^{-1}$ (un-purl).
    \item \textbf{Identity:} The empty string $\epsilon$.
\end{itemize}

\paragraph{Operational Logic}
The operation is string concatenation followed by recursive reduction. The reduction rule is defined by the cancellation of adjacent inverse pairs:
\begin{equation}
    x x^{-1} \to \epsilon \quad \text{and} \quad x^{-1} x \to \epsilon
\end{equation}
Specifically, the pairs $(k, K)$ and $(p, P)$ annihilate each other. This process repeats until the string is in its irreducible canonical form.

\paragraph{Example}
Computing $\texttt{kp} \gg \texttt{PK}$:
\begin{enumerate}
    \item Concatenate: $\texttt{kpPK}$
    \item Inner Cancellation: The adjacent pair $p, P$ cancels. $\texttt{k(pP)K} \to \texttt{kK}$.
    \item Secondary Cancellation: The adjacent pair $k, K$ cancels. $\texttt{kK} \to \epsilon$.
    \item Result: Empty string (Identity).
\end{enumerate}

\subsection{Rubik's Cube Sequence System (Finite Non-Abelian Group)}
This domain models the algebra of face rotations on a $3\times3\times3$ cube. It enforces strict canonicalization rules involving commutation and modulo arithmetic, representing a quotient of a free group.

\paragraph{Representation}
\begin{itemize}
    \item \textbf{Moves:} $F \in \{R, L, U, D, F, B\}$ representing face turns.
    \item \textbf{Modifiers:} None (90° clockwise), $\#$ (90° counter-clockwise/prime), $2$ (180°).
    \item \textbf{Sequence:} A space-separated list of moves (e.g., \texttt{R U R\#}).
\end{itemize}

\paragraph{Operational Logic}
Operations are defined by sequence concatenation followed by reduction to a unique \textbf{Canonical Form}. Three specific rules are applied iteratively:
\begin{enumerate}
    \item \textbf{Inverse Cancellation:} Moves followed by their inverse are removed (e.g., $R R\# \to \epsilon$).
    \item \textbf{Face Consolidation (Mod 4):} Consecutive moves of the same face are summed modulo 4.
    \begin{itemize}
        \item Example: $R \cdot R2 \to R3 \equiv R\#$ (since $1+2=3 \equiv -1$).
        \item Example: $R \cdot R \cdot R \cdot R \to R4 \equiv \epsilon$.
    \end{itemize}
    \item \textbf{Commutative Reordering:} While the group is generally non-abelian, opposite faces commute (e.g., $RL = LR$). To ensure a unique representation, we enforce a lexicographical priority:
    \begin{itemize}
        \item $R$ precedes $L$; $U$ precedes $D$; $F$ precedes $B$.
        \item Example: A sequence input as $\texttt{L R}$ is automatically rewritten to $\texttt{R L}$.
    \end{itemize}
\end{enumerate}

\paragraph{Example}
Solving equation $X = \text{inv}(\texttt{R U}) \cdot \texttt{R U R\#}$:
\begin{enumerate}
    \item Invert $\texttt{R U} \to \texttt{U\# R\#}$.
    \item Concatenate: $\texttt{U\# R\# R U R\#}$.
    \item Consolidate: $R\#$ and $R$ cancel. Sequence becomes $\texttt{U\# U R\#}$.
    \item Consolidate: $U\#$ and $U$ cancel.
    \item Result: $\texttt{R\#}$.
\end{enumerate}

\end{document}